\documentclass[11pt]{article}

\usepackage[preprint]{acl}

\usepackage{times}
\usepackage{latexsym}
\usepackage{amsmath} 
\usepackage{amssymb}
\usepackage[T1]{fontenc}
\usepackage[utf8]{inputenc}
\usepackage{microtype}
\usepackage{inconsolata}
\usepackage{verbatim}
\usepackage{tcolorbox}
\tcbuselibrary{breakable}

\usepackage{amsmath,amssymb}
\usepackage{algorithm}
\usepackage{algorithmic}
\usepackage{tikz}
\usepackage{xcolor}
\definecolor{mygreen}{HTML}{008000} 
\definecolor{myred}{HTML}{B22222}

\usetikzlibrary{arrows.meta,positioning,calc,decorations.pathreplacing}

\definecolor{lightred}{rgb}{1, 0.8, 0.8}  
\definecolor{lightgreen}{rgb}{0.77, 1, 0.77}  

\usepackage{tikz}
\usetikzlibrary{positioning,arrows.meta}
\usepackage{pgfplots}
\pgfplotsset{compat=1.18}

\definecolor{advpos}{RGB}{220, 255, 220}   
\definecolor{advneg}{RGB}{255, 220, 220}   
\definecolor{advneu}{RGB}{240, 240, 240}   
\definecolor{bordercol}{RGB}{80, 80, 80}   

\usepackage{graphicx}
\usepackage{booktabs}
\usepackage{multirow}
\usepackage{colortbl} 
\usepackage{xcolor}
\usepackage{enumitem}

\definecolor{Gray}{gray}{0.9}

\usepackage{tikz}
\usetikzlibrary{shapes, arrows.meta, positioning, calc, backgrounds, fit, decorations.pathreplacing, shadows}
\definecolor{tape_gray}{RGB}{240, 240, 240}
\definecolor{node_white}{RGB}{255, 255, 255}
\definecolor{myblue}{RGB}{220, 240, 255}
\definecolor{mygreen}{RGB}{220, 255, 220}

\usepackage{tikz}
\usetikzlibrary{arrows.meta,positioning,calc,fit,decorations.pathreplacing}

\definecolor{bad}{RGB}{255, 200, 200}   
\definecolor{good}{RGB}{200, 255, 200}  
\definecolor{neutral}{RGB}{240, 240, 240} 

\usepackage{xspace}
\newcommand{\ourmethod}{TreeAdv\xspace}
\newcommand{\ourmetho}{TreeAdv}

\title{TreeAdv: Tree-Structured Advantage Redistribution for Group-Based RL}

\author{
    \begin{tabular}{c}
        \textbf{Lang Cao\textsuperscript{*} \quad Hui Ruan\textsuperscript{*} \quad Yongqian Li\textsuperscript{*} \quad Chao Peng} \\        
        \textbf{Wu Ning \quad Haonan Song \quad Renhong Chen\textsuperscript{\dag} \quad Yitong Li\textsuperscript{\dag}} \\        
        \small{Huawei Technologies Co., Ltd., China}\\
        \small{\texttt{\{caolang4,liyitong3\}@huawei.com}}
    \end{tabular}
}

\begin{document}
\maketitle


\begin{abstract}
Reinforcement learning with group-based objectives, such as Group Relative Policy Optimization (GRPO), is a common framework for aligning large language models on complex reasoning tasks.
However, standard GRPO treats each rollout trajectory as an independent sequence and assigns one single sequence-level advantage to all tokens, which leads to sample inefficiency and a length bias toward verbose or redundant chains of thought without improving logical depth.
We introduce \ourmethod (Tree-Structured Advantage Redistribution for Group-Based RL), which makes the tree structure of group rollouts explicit for both exploration and advantage assignment.
Specifically, \ourmethod builds a group of trees (a forest) based on an entropy-driven sampling method where each tree branches at high-uncertainty decisions while sharing low-uncertainty tokens across rollouts.
Then, \ourmethod aggregates token-level advantages for internal tree segments by redistributing the advantages of complete rollouts (all leaf nodes), and \ourmethod can easily apply to group-based objectives such as GRPO or GSPO.
Across $10$ math reasoning benchmarks, \ourmethod consistently outperforms GRPO and GSPO, while using substantially fewer generated tokens under identical supervision, data, and decoding budgets.
\end{abstract}

\section{Introduction}

Since the emergence of o1-style reasoning models~\cite{openai2024openaio1card,o1_preview_openai_2024_en}, test-time scaling~\cite{muennighoff2025s1simpletesttimescaling} has received renewed attention: allocating more computation at inference—e.g., spending more time “thinking” or exploring multiple candidate reasoning trajectories—can significantly improve performance on difficult reasoning tasks~\cite{jiang2025thinkneedlargehybridreasoning,gemini3pro_modelcard_2025}. 

To fully benefit from test-time scaling, models must be trained to produce reliable multi-step reasoning trajectories rather than merely longer ones. Reinforcement learning (RL) post-training is an effective mechanism for eliciting reliable multi-step reasoning in large language models (LLMs), especially for mathematics problem solving~\cite{zhang2025surveyreinforcementlearninglarge,phan2025humanitysexam}.
Group-based policy optimization methods such as GRPO~\cite{shao2024deepseekmathpushinglimitsmathematical} and GSPO~\cite{zheng2025groupsequencepolicyoptimization} are increasingly adopted as practical \emph{PPO-style} recipes for scaling reasoning RL: they use group statistics as baselines to avoid training a separate value network, reducing memory overhead while retaining PPO-like policy updates~\cite{yang2025qwen3technicalreport,deepseekai2025deepseekr1incentivizingreasoningcapability}.

However, GRPO-style training largely relies on \emph{sequence-level credit assignment}~\cite{li2025saltstepleveladvantageassignment}, where each sampled response $y$ receives a terminal scalar reward, which is converted into a single advantage $A$ and applied uniformly to all tokens of $y$.
This can reinforce redundant or locally suboptimal reasoning whenever the final answer is correct and yields noisy updates because key decisions, corrected mistakes, and detours share the same scalar supervision~\cite{ye2025correctnessharmonizingprocessoutcome,zhang2025surveyreinforcementlearninglarge}.

We address this limitation without dense process supervision or additional reward models.
Instead, we exploit the shared-prefix structure among multiple sampled continuations from the same prompt, where these continuations naturally form a tree of branched rollouts.
By aggregating terminal results over this topology, we derive structure-aware segment or token advantages for intermediate decisions~\cite{yang2025treerpotreerelativepolicy}.

We propose Tree-Structured Advantage Redistribution for Group-Based RL (\ourmethod), a drop-in modification to PPO-style group-based RL (GRPO/GSPO).
Empirically, \ourmethod improves long-form reasoning performance and reduces the average generated tokens under matched rollout budgets and decoding settings.
In more detail, \ourmethod calculate token-level advantages by re-organizing rollout trajectories which are sampled by a designed entropy based branching strategy.
And then \ourmethod utilize these token-level advantages for any group based RL methods, such as GRPO and GSPO.
We demonstrate our method over commonly used math reasoning benchmarks and compared with sota baselines, where our method consistently outperforms them in both accuracy and token-consumption.

Our contributions are: (i) topology-aware advantage redistribution for GRPO/GSPO with minimal pipeline changes; (ii) improved accuracy and shorter reasoning on long-form math benchmarks; (iii) analysis of sequence-level credit assignment failure modes and their mitigation.

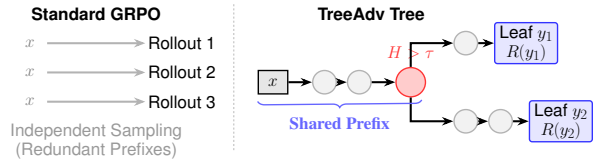
\begin{figure}[!t]
    \centering
    \resizebox{\columnwidth}{!}{%
    \begin{tikzpicture}[
        font=\sffamily\Huge,
        >=Stealth,
        token/.style={circle, draw=gray!60, fill=gray!10, inner sep=0pt, minimum size=1.2cm, line width=2pt},
        highlight/.style={circle, draw=red!80, fill=red!20, thick, inner sep=0pt, minimum size=1.5cm, line width=2pt},
        leaf/.style={rectangle, draw=blue!80, fill=blue!10, rounded corners, minimum height=1.8cm, minimum width=3.2cm, align=center, line width=2pt},
        title_text/.style={font=\sffamily\bfseries\Huge, align=center}
    ]

    \node[title_text] at (3, 3) {Standard GRPO};
    
    \node[anchor=east, gray] at (0, 1.5) {$x$};
    \draw[->, line width=3pt, gray!60] (0.5, 1.5) -- (5.5, 1.5);
    \node[right] at (5.5, 1.5) {Rollout 1};

    \node[anchor=east, gray] at (0, 0) {$x$};
    \draw[->, line width=3pt, gray!60] (0.5, 0) -- (5.5, 0);
    \node[right] at (5.5, 0) {Rollout 2};

    \node[anchor=east, gray] at (0, -1.5) {$x$};
    \draw[->, line width=3pt, gray!60] (0.5, -1.5) -- (5.5, -1.5);
    \node[right] at (5.5, -1.5) {Rollout 3};
    
    \node[gray!80, align=center] at (3, -3.5) {Independent Sampling\\(Redundant Prefixes)};

    \draw[dashed, gray!40, line width=2pt] (10, 3.5) -- (10, -4);

    \def\offset{12} 
    
    \node[title_text] at (\offset + 5, 3) {\ourmethod Tree};

    \node[draw, rectangle, thick, fill=gray!20, minimum width=1.5cm, minimum height=1.2cm, line width=2pt] (root) at (\offset, -0.5) {$x$};

    \node[token, right=1.2cm of root] (t1) {};
    \node[token, right=0.5cm of t1] (t2) {};
    
    \node[highlight, right=1.2cm of t2] (split) {};
    \node[above=0.3cm of split, text=red!80, font=\bfseries\Huge] {$H > \tau$};
    
    \node[token, above right=1.0cm and 1.8cm of split] (b1) {};
    \node[leaf, right=0.8cm of b1] (leaf1) {Leaf $y_1$\\$R(y_1)$};
    
    \node[token, below right=1.0cm and 1.8cm of split] (b2_1) {};
    \node[token, right=0.5cm of b2_1] (b2_2) {};
    \node[leaf, right=0.8cm of b2_2] (leaf2) {Leaf $y_2$\\$R(y_2)$};

    \draw[->, line width=3pt] (root) -- (t1);
    \draw[->, line width=3pt] (t1) -- (t2);
    \draw[->, line width=3pt] (t2) -- (split);
    \draw[->, line width=3pt] (split) |- (b1);
    \draw[->, line width=3pt] (b1) -- (leaf1);
    \draw[->, line width=3pt] (split) |- (b2_1);
    \draw[->, line width=3pt] (b2_1) -- (b2_2);
    \draw[->, line width=3pt] (b2_2) -- (leaf2);

    \draw[decorate, decoration={brace, amplitude=10pt, mirror, raise=10pt}, blue!60, line width=2pt] 
        (root.south west) -- (split.south east) 
        node[midway, yshift=-1.5cm, text=blue!60, font=\bfseries\Huge] {Shared Prefix};

    \end{tikzpicture}%
    }
    \caption{Comparison of trajectory generation. \textbf{Left:} Standard GRPO samples independent rollouts that redundantly repeat prefixes. \textbf{Right:} \ourmethod constructs a tree by sharing prefixes and branching only at high-entropy states ($H > \tau$).}
    \label{fig:smart_overview}
\end{figure}

\section{Related Work}

\subsection{Group-based Policy Optimization}

PPO-style policy optimization~\cite{schulman2017proximalpolicyoptimizationalgorithms} with GAE-style estimators~\cite{schulman2018highdimensionalcontinuouscontrolusing} is effective but often costly for LLM post-training due to the value network.
Group-based policy optimization~\cite{chen2025heterogeneousgroupbasedreinforcementlearning,nimmaturi2025predictivescalinglawsefficient} replaces learned baselines with group statistics, enabling PPO-style updates without training a separate critic/value network.
GRPO~\cite{shao2024deepseekmathpushinglimitsmathematical} computes group-normalized advantages from sampled rollouts, and GSPO~\cite{zheng2025groupsequencepolicyoptimization} improves stability via sequence-level smoothing/weighting while preserving the same PPO-style training backbone~\cite{yang2025qwen3technicalreport}.
A shared limitation is sequence-level supervision: outcomes are defined on complete rollouts, making it difficult to assign distinct credit to intermediate reasoning segments~\cite{li2025saltstepleveladvantageassignment,ye2025correctnessharmonizingprocessoutcome}.

\subsection{Tree-based Reasoning and Rollout Strategies}

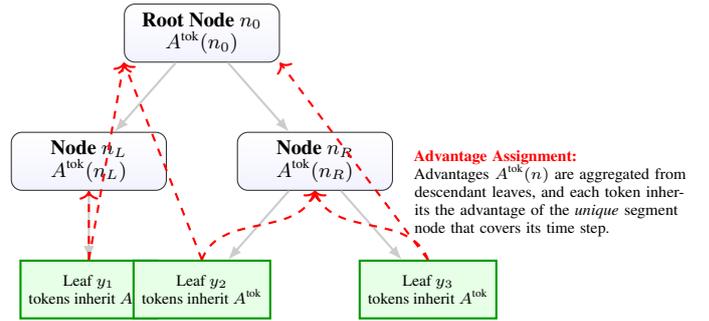
\begin{figure}[t]
    \centering
    \begin{tikzpicture}[
        scale=0.85, transform shape,
        level distance=2.0cm,
        sibling distance=3.5cm,
        edge from parent/.style={draw, thick, ->, >=latex, black!20},
        segment/.style={rectangle, draw=black!80, top color=white, bottom color=blue!5, rounded corners, minimum height=0.9cm, minimum width=2.4cm, align=center, font=\footnotesize},
        leaf/.style={rectangle, draw=green!60!black, fill=green!10, thick, minimum height=0.9cm, minimum width=2.0cm, align=center, font=\scriptsize},
        mathlabel/.style={midway, fill=white, font=\tiny, text=gray}
    ]

    \node[segment] (seg_root) {\textbf{Root Node $n_0$}\\$A^{\text{tok}}(n_0)$}
        child {
            node[segment] (seg_left) {\textbf{Node $n_L$}\\$A^{\text{tok}}(n_L)$}
            child {
                node[leaf] (leaf1) {Leaf $y_1$\\tokens inherit $A^{\text{tok}}$}
                edge from parent node[left, font=\tiny] {}
            }
        }
        child {
            node[segment] (seg_right) {\textbf{Node $n_R$}\\$A^{\text{tok}}(n_R)$}
            child {
                node[leaf] (leaf2) {Leaf $y_2$\\tokens inherit $A^{\text{tok}}$}
                edge from parent node[right, font=\tiny] {}
            }
            child {
                node[leaf] (leaf3) {Leaf $y_3$\\tokens inherit $A^{\text{tok}}$}
                edge from parent node[right, font=\tiny] {}
            }
        };

    \draw[->, red, dashed, thick] (leaf1.north) -- (seg_left.south);
    \draw[->, red, dashed, thick] (leaf2.north) to[out=70, in=-110] (seg_right.south);
    \draw[->, red, dashed, thick] (leaf3.north) to[out=110, in=-70] (seg_right.south);
    
    \draw[->, red, dashed, thick] (leaf1.north) -- (seg_root.south west);
    \draw[->, red, dashed, thick] (leaf2.north) -- (seg_root.south west);
    \draw[->, red, dashed, thick] (leaf3.north) -- (seg_root.south east);

    \node[right=0.2cm of seg_right, align=left, font=\scriptsize, text width=4.2cm, yshift=-0.5cm] {
        \textbf{\color{red} Advantage Assignment:}\\
        Advantages $A^{\text{tok}}(n)$ are aggregated from descendant leaves, and each token
        inherits the advantage of the \emph{unique} segment node that covers its time step.\\[0.2cm]
    };

    \end{tikzpicture}
    \caption{Illustration of \ourmetho's tree-aware advantage construction. Scalar rewards are computed only on complete rollouts (leaves) and converted to sequence-level advantages. These leaf advantages are aggregated into segment-level advantages $A^{\text{tok}}(n)$. Each token then receives the advantage of the unique segment node that covers its time step.}
    \label{fig:smart_mechanism}
\end{figure}

Tree-structured search improves inference-time reasoning by exploring branched continuations, as in ToT~\cite{yao2023treethoughtsdeliberateproblem} and MCTS-style methods~\cite{ding2024thoughtsdefyinglawpenrose}, with strong results on challenging benchmarks~\cite{wang2025mctsjudgetesttimescalingllmasajudge,zhang2024restmctsllmselftrainingprocess}.
For training, TreeRL~\cite{hou2025treerlllmreinforcementlearning} derives step-level advantages via Monte Carlo value/return estimation over descendant outcomes, while TreeRPO~\cite{yang2025treerpotreerelativepolicy} performs local sibling normalization at each step.
TreePO~\cite{li2025treepobridginggappolicy} adopts DAPO~\cite{yu2025dapoopensourcellmreinforcement} for process preference optimization, which in their setting requires additional rollouts (or extra sampled candidates) to construct per-step preference comparisons, along with extra per-step scoring and bookkeeping.
Differently, our method keeps supervision terminal and critic-free, and uses shared-prefix topology to redistribute terminal group-relative advantages to internal segments via subtree-based aggregation(See figure~\ref{fig:smart_mechanism}).

\section{Method}
\label{sec:method}

\subsection{Motivation}
\label{sec:motivation}

A recurring challenge in RL post-training for multi-step reasoning is that not all tokens are equally decision-critical~\cite{wang2024chainofthoughtreasoningprompting}.
In practice, models often exhibit low uncertainty for most local steps, while a small number of positions can concentrate higher ambiguity: different plausible next tokens can lead to substantially different downstream reasoning trajectories~\cite{ma2025estimatingllmuncertaintyevidence}.
This suggests that both exploration and credit assignment should focus on these high-uncertainty positions, rather than treating every token in a rollout as equally informative.

Prior work has explored allocating branching or process-level effort based on token uncertainty (e.g., entropy)~\citep{li2025treepobridginggappolicy,cao2026bangbuckprocessreward}.
The key intuition is simple: branching at high-entropy positions produces rollouts that differ in meaningful local decisions under the same sampling budget, making the resulting outcomes more informative for learning.

These observations naturally connect to group-based RL.
If we branch at uncertain prefixes, the sampled continuations form a shared-prefix tree, where leaves correspond to complete rollouts with terminal rewards.
This structure enables prefix-conditioned comparisons among alternative local decisions made at the same prefix.
\ourmethod leverages this topology to redistribute terminal, group-normalized advantages from leaf rollouts back to the corresponding shared-prefix segments, yielding token-/segment-level learning signals while keeping the standard clipped surrogate objective unchanged.

\subsection{Background: Group-Based RL}
\label{sec:prelim}

For a prompt $x$, we sample a group of $K$ rollout trajectories, where the $i$-th trajectory at step $t$ is denoted as $s_{i, t}=(x, a_{i, 1}, \cdots, a_{i,t})$ from a policy $\pi_{\theta}(a_{i,t}\mid s_{i,<t})$.
Group-based objectives (e.g., GRPO) rely solely on a scalar sequence-level reward per rollout, $R_i$, and form a group-normalized sequence advantage without a value network:
\begin{equation}
A_i=\frac{R_i-\mu_R}{\sigma_R} \, , 
\label{eq:grpo_seq_adv}
\end{equation}
where $\mu_R =\frac{1}{K}\sum_{j=1}^K R_j$, and $\sigma_R =\sqrt{\frac{1}{K}\sum_{j=1}^K (R_j-\mu_R)^2+\delta}$.

In the following, we take GRPO as an example of applying a token-level optimization using our \ourmethod.
Let $\pi_{\text{old}}$ denote the behavior policy for generating rollouts.
GRPO optimizes a PPO style clipped surrogate with per-token likelihood ratios $r_{i,t}(\theta)=\frac{\pi_{\theta}(a_{i,t}\mid s_{i,<t})}{\pi_{\text{old}}(a_{i,t}\mid s_{i,<t})}$, and $
\bar r_{i,t}(\theta)=\operatorname{clip}\!\left(r_{i,t}(\theta),1-\epsilon,1+\epsilon\right)$, and the final objective $\mathcal{L}_{\text{GRPO}}(\theta)$ is defined as
\begin{align}\label{eq:grpo_obj}
\mathbb{E}\left[
\frac{1}{K}\sum_{i=1}^{K}\sum_{t=1}^{|s_i|}
\min\!\left(
r_{i,t}(\theta)A_i,
\bar r_{i,t}(\theta)A_i
\right)
\right]
\end{align}
In this setup, the advantage score $A_i$ is applied at the sequence level and to all tokens within the rollout $s_i$.

\subsection{\ourmetho: Entropy-Guided Trees and Token-Level Advantage}
\label{sec:smart_method}

Recent studies show that tokens vary in their importance and should consequently receive distinct advantage scores~ \cite{wang2024mathshepherdverifyreinforcellms,yang2025treerpotreerelativepolicy}.
Empirically, we observed that most of the reasoning steps generated by current RL models are correct, while a few intermediates of them make inference errors.
These critical reasoning steps should be reinforced more, while standard group-based RL fails to distinguish them, leading to sample inefficiency and a tendency to generate unnecessarily long rollouts.

Motivated by these observations, we proposed a method, \ourmethod, which considers token-level advantages.
Specifically, our method samples rollout trees (rather than independent sequences) using an entropy-driven branching strategy.
Then, \ourmethod converts leaf (complete-rollout) advantages into token-level advantages by organizing the $K$ rollouts of each prompt into a prefix forest and aggregating advantages on shared prefix segments.

\paragraph{Entropy-guided branching.}
We measure uncertainty at a prefix state $s_t$, omitting the index $i$, by the token entropy under the behavior policy:
\begin{equation}
H(s_t)=-\sum_{a\in\mathcal{V}}\pi_{\text{old}}(a_t \mid s_{<t})\log \pi_{\text{old}}(a_t \mid s_{<t}). \nonumber
\end{equation}
Starting from the prompt, we perform linear rollouts under $\pi_{\text{old}}$.
Positions where entropy $H(s_t)$ exceeds a threshold $\tau$ are treated as \emph{branching points}. At these points, we sample multiple distinct next tokens to create child nodes, expanding the frontier (Appendix~\ref{app:entropy}).
This process yields (i) a set of completed rollouts (leaves) for reward evaluation, and (ii) internal nodes representing shared reasoning segments.

\paragraph{Assigning token advantages}
For each rollout trajectory $s_i$, we compute the standard group-normalized sequence advantages $A_i$ as in Eq.~\ref{eq:grpo_seq_adv}.
The token advantage $A_{i,t}$ of token $a_{i,t}$ is defined by summing up the advantage scores of the trajectory set $S(a_{i,t})$, consisting of all trajectories sharing $a_{i,t}$, and then normalized by the number of trajectories,
\begin{equation}
A^{\text{tok}}_{i,t}
=
\frac{1}{\left| S(a_{i,t}) \right|}
\sum_{\ell\in S(a_{i,t})} A_{\ell}.
\label{eq:token_adv}
\end{equation}
The normalize operation can prevent advantages near the root (with many descendants) from dominating by scale.

\paragraph{\ourmethod objective in GRPO}
\ourmethod integrates $\tilde A^{\text{tok}}_{i,t}$ into the same clipped surrogate as GRPO, the objective $\mathcal{L}_{\text{\ourmetho}}(\theta)$ is defined as,
\begin{equation}\label{eq:smart_obj}
\mathbb{E}\left[
\frac{1}{K}\sum_{i=1}^{K}\sum_{t=1}^{|s_i|}
\min\!\left(
r_{i,t}(\theta)\tilde A^{\text{tok}}_{i,t},
\bar r_{i,t}(\theta)\tilde A^{\text{tok}}_{i,t}
\right)
\right] 
\end{equation}
All other components, e.g., KL regularization to a reference policy, remain unchanged, and \ourmethod only modifies the construction of advantages on the token level.

\paragraph{\ourmethod objective in GSPO}
\ourmethod integrates $\tilde A^{\text{tok}}_{i,t}$ into the same clipped surrogate form, while replacing a length-normalized sequence-level importance ratio (PPO-style) with the per-token likelihood ratio.
We follow the definition of GSPO-token and define the length-normalized sequence ratio for trajectory $i$ as
\begin{equation}\label{eq:gspo_seq_ratio}
w_i(\theta)
=
\exp\left(
\frac{1}{|s_i|}
\sum_{t=1}^{|s_i|}
\log
\frac{
\pi_{\theta}(a_{i,t}\mid s_{i,<t})
}{
\pi_{\text{old}}(a_{i,t}\mid s_{i,<t})
}
\right)
\end{equation}
and construct a token-wise ratio
\begin{equation}
w_{i,t}(\theta)
=
\operatorname{sg}\!\left[w_i(\theta)\right]\cdot
\frac{
\pi_{\theta}(a_{i,t}\mid s_{i,<t})
}{
\operatorname{sg}\!\left[\pi_{\theta}(a_{i,t}\mid s_{i,<t})\right]
},
\label{eq:gspo_token_inject}
\end{equation}
where $\operatorname{sg}[\cdot]$ denotes the stop-gradient operator.
We further define the clipped ratio $\bar w_{i,t}(\theta) = \operatorname{clip}\!\left(w_{i,t}(\theta),1-\epsilon,1+\epsilon\right)$.
Then, the objective $\mathcal{L}_{\text{\ourmetho}}(\theta)$ for GSPO is defined as
\begin{equation}
\mathbb{E}\left[
\frac{1}{K}\sum_{i=1}^{K}\sum_{t=1}^{|s_i|}
\min\!\left(
w_{i,t}(\theta)\tilde A^{\text{tok}}_{i,t},
\bar w_{i,t}(\theta)\tilde A^{\text{tok}}_{i,t}
\right)
\right].
\label{eq:smart_gspo_obj}
\end{equation}
Similarly, \ourmethod only modifies the construction of advantages on the token level while adopting a GSPO-style sequence ratio for clipping and optimization.

\begin{table*}[!t]
\centering
\small
\setlength{\tabcolsep}{3.2pt}
\renewcommand{\arraystretch}{1.06}

\newcommand{\mindent}{\hspace{0.6em}}
\newcommand{\sindent}{\hspace{1.2em}}


\resizebox{\textwidth}{!}{%
\begin{tabular}{@{}lcccccccccccc@{}}
\toprule
\multirow{2}{*}{\textbf{Method}}
& \textbf{AIME24} & \textbf{AIME25} & \textbf{BRUMO} & \textbf{CMI}
& \textbf{GPQA} & \textbf{HMMT} & \textbf{MATH}
& \textbf{OlymE} & \textbf{OlymH} & \textbf{Omni}
& \textbf{Avg} $\uparrow$ & \textbf{Tok} $\downarrow$ \\
& {\scriptsize(32p1)} & {\scriptsize(32p1)} & {\scriptsize(32p1)} & {\scriptsize(32p1)}
&  & {\scriptsize(32p1)} & 
&  &  & {\scriptsize}
& {\scriptsize(\%)} & {\scriptsize(\#)} \\
\midrule

\multicolumn{13}{@{}l@{}}{\textit{\textbf{(A) SOTA open models (DeepSeek)}}} \\
\addlinespace[0.15em]
\mindent \texttt{DeepSeek-R1-Distill-Qwen-7B} & 53.23 & 38.96 & 51.67 & 28.98 & 23.74 & 24.48 & 91.60 & 42.00 & 14.00 & 44.26 & 41.29 & 12222 \\
\mindent \texttt{DeepSeek-R1-Distill-Qwen-14B} & 69.90 & 48.75 & 60.21 & 36.80 & 41.92 & 33.33 & 92.80 & 51.00 & 17.00 & 49.55 & 50.13 & 12118 \\
\mindent \texttt{DeepSeek-R1-Distill-Qwen-32B} & 69.06 & 54.38 & 64.38 & 40.55 & 40.91 & 33.33 & 94.00 & 64.00 & 13.00 & 51.60 & 52.52 & 11603 \\
\mindent \texttt{DeepSeek-R1-Distill-Llama-8B} & 41.98 & 30.31 & 38.02 & 19.84 & 29.29 & 21.88 & 81.00 & 29.00 & 7.00 & 38.03 & 33.64 & 12854 \\

\cmidrule(lr){1-13}
\multicolumn{13}{@{}l@{}}{\textit{\textbf{(B) Our \ourmethod on base models}}} \\
\addlinespace[0.10em]
\sindent \texttt{Qwen3-8B-Base} & 11.25 & 9.48 & 18.85 & 4.45 & 18.69 & 1.35 & 68.20 & 7.00 & 3.00 & 24.07 & 16.64 & 2667 \\
\sindent \texttt{ w. GRPO} & 30.00 & 22.00 & 37.00 & 16.00 & 48.00 & 12.00 & 88.00 & 17.00 & 4.00 & 36.00 & 31.00 & 7025 \\
\sindent \texttt{ w. \ourmetho-GRPO} & 27.71 & 24.79 & 37.19 & 16.09 & 49.49 & 11.77 & 85.60 & 20.00 & 10.00 & 34.03 & \colorbox{lightgreen}{31.67} & 6171 \\
\sindent \texttt{ w. GSPO} & 22.08 & 17.40 & 32.40 & 11.33 & 41.41 & 9.38 & 84.40 & 6.00 & 2.00 & 31.91 & 25.83 & 9583 \\
\sindent \texttt{ w. \ourmetho-GSPO} & 21.98 & 16.35 & 29.79 & 11.41 & 42.42 & 8.02 & 85.00 & 11.00 & 6.00 & 30.71 & \colorbox{lightgreen}{26.27} & 8319 \\

\cmidrule(lr){2-13}
\sindent \texttt{Qwen3-4B-Base} & 9.58 & 8.13 & 18.33 & 3.52 & 24.75 & 1.35 & 71.60 & 6.00 & 1.00 & 24.57 & 16.88 & 2628 \\
\sindent \texttt{ w. GRPO} & 21.88 & 13.33 & 24.69 & 8.05 & 40.91 & 5.00 & 79.60 & 10.00 & 1.00 & 28.70 & 23.32 & 2794 \\
\sindent \texttt{ w. \ourmetho-GRPO} & 21.25 & 19.17 & 23.02 & 12.11 & 45.45 & 8.85 & 84.00 & 15.00 & 3.00 & 32.05 & \colorbox{lightgreen}{26.39} & 5894 \\
\sindent \texttt{ w. GSPO} & 14.58 & 10.52 & 21.15 & 5.31 & 42.42 & 1.46 & 78.00 & 9.00 & 2.00 & 25.43 & 20.99 & 6346 \\
\sindent \texttt{ w. \ourmetho-GSPO} & 13.44 & 12.29 & 19.48 & 7.03 & 40.91 & 3.13 & 78.60 & 9.00 & 4.00 & 25.34 & \colorbox{lightgreen}{21.32} & 8606 \\

\cmidrule(lr){1-13}
\multicolumn{13}{@{}l@{}}{\textit{\textbf{(C) Our \ourmethod on instruct models}}} \\

\sindent \texttt{Qwen3-4B-Inst Think} & 71.98 & 64.90 & 63.23 & 43.91 & 54.04 & 42.08 & 94.20 & 67.00 & 20.00 & 53.16 & 57.45 & 15322 \\
\sindent \texttt{ w. GRPO} & 70.94 & 62.50 & 60.83 & 40.47 & 54.04 & 38.02 & 94.40 & 68.00 & 16.00 & 52.30 & 55.75 & 15000 \\
\sindent \texttt{ w. \ourmetho-GRPO} & 71.88  & 63.96 & 61.15 & 43.36 & 53.03 & 41.25 & 93.40 & 67.00 & 19.00 & 52.55 & \colorbox{lightred}{56.66} & 13439 \\

\cmidrule(lr){2-13}
\sindent \texttt{Qwen3-4B-Inst Non-Think} & 22.08 & 18.65 & 31.15 & 15.94 & 40.91 & 11.04 & 83.40 & 16.00 & 4.00 & 33.83 & 27.70 & 3333 \\ 
\sindent \texttt{ w. GRPO} & 46.04 & 34.48 & 47.92 & 20.70 & 48.48 & 21.15 & 89.00 & 36.00 & 9.00 & 43.29 & 39.61 & 9710 \\
\sindent \texttt{ w. \ourmetho-GRPO} & 45.00 & 39.69 & 47.81 & 25.86 & 49.49 & 25.21 & 89.60 & 37.00 & 9.00 & 43.70 & \colorbox{lightgreen}{41.24} & 8015 \\

\cmidrule(lr){2-13}
\sindent \texttt{Qwen3-8B-Inst Think} & 74.58 & 66.35 & 67.40 & 45.47 & 61.62 & 40.94 & 93.40 & 74.00 & 18.00 & 54.22 & 59.60 & 16200 \\
\sindent \texttt{ w. GRPO} & 75.21 & 66.77 & 67.71 & 45.16 & 58.08 & 43.96 & 93.60 & \textbf{78.00} & 23.00 & 53.97 & 60.55 & 15693 \\
\sindent \texttt{ w. \ourmetho-GRPO} & 75.83 & \textbf{68.13} & 69.27 & 46.64 & 61.62 & \textbf{45.31} & 93.80 & \textbf{78.00} & \textbf{27.00} & 54.34 & \colorbox{lightgreen}{\textbf{61.99}} & 12073 \\

\cmidrule(lr){2-13}
\sindent \texttt{Qwen3-8B-Inst Non-Think} & 28.02 & 21.15 & 30.94 & 14.69 & 44.44 & 10.52 & 83.20 & 23.00 & 4.00 & 35.41 & 29.54 & 3506 \\
\sindent \texttt{ w. GRPO} & 56.25 & 40.00 & 52.92 & 30.31 & 59.60 & 26.98 & 90.80 & 44.00 & 11.00 & 50.52 & {46.24} & {13373} \\
\sindent \texttt{ w. \ourmetho-GRPO} & 60.63 & 49.17 & 55.00 & 33.44 & 54.55 & 29.90 & 92.20 & 58.00 & 12.00 & 48.83 & \colorbox{lightgreen}{49.37} & 9962 \\
\sindent \texttt{ w. GSPO} & 58.02 & 46.98 & 55.10 & 28.13 & 55.05 & 28.96 & 93.00 & 48.00 & 11.00 & 48.13 & 47.24 &  9420\\
\sindent \texttt{ w. \ourmetho-GSPO} & 60.83 & 48.33 & 55.21 & 30.63 & 57.58 & 29.58 & 91.60 & 46.00 & 14.00 & 47.74 & 48.15 & 11750 \\

\cmidrule(lr){2-13}
\sindent \texttt{Qwen3-4B-Inst-2507} & 61.56 & 45.73 & 57.81 & 32.03 & 60.10 & 30.31 & 93.40 & 57.00 & 11.00 & 49.55 & 49.85 & 7701 \\
\sindent \texttt{ w. GRPO}  & 60.73 & 46.25 & 54.58 & 30.86 & 60.61 & 29.69 & 93.20 & 54.00 & 9.00 & 48.40 & 48.73 & 10397 \\
\sindent \texttt{ w. \ourmetho-GRPO} & 61.98 & 52.81 & 56.98 & 32.34 & 64.14 & 31.67 & 93.80 & 53.00 & 18.00 & 50.72 & \colorbox{lightgreen}{51.54} & 5805 \\
\sindent \texttt{ w. GSPO}  & 60.10 & 45.10 & 54.90 & 31.56 & 60.10 & 29.69 & 92.69 & 59.00 & 11.00 & 50.34 &  49.44 & 6332 \\
\sindent \texttt{ w. \ourmetho-GSPO} & 63.44 & 53.65 & 58.85 & 32.34 & 63.13 & 33.85 & 93.60 & 54.00 & 13.00 & 50.09 & \colorbox{lightgreen}{51.60} & 6197 \\

\cmidrule(lr){2-13}
\sindent \texttt{Qwen3-30B-A3B-2507} & 74.48 & 61.15 & \textbf{72.08} & 44.38 & 67.17 & 44.48 & 94.60 & 73.00 & 20.00 & \textbf{55.96} & 60.73 & 6291 \\
\sindent  \texttt{ w. GSPO} & 63.00 & 60.63 & 68.02 & 45.08 & \textbf{71.72} & 42.60 & 95.60 & 76.00 & 25.00 & 55.44 & 60.31 & 5633 \\
\sindent \texttt{  w. \ourmetho-GSPO} & \textbf{75.63} & 62.71 & 68.90 & \textbf{47.11} & 68.18 & 43.54 & \textbf{95.60} & 76.00 & 24.00 & 55.76 & \colorbox{lightgreen}{61.75} & 5748 \\
\bottomrule
\end{tabular}
}
\caption{
Results on a suite of reasoning benchmarks. (A) reports representative open SOTA models (DeepSeek family). (B) reports our \ourmethod on base models. (C)  evaluates our \ourmethod on instruct models, using GRPO/GSPO as the corresponding baselines. For all \ourmethod entries in this table, we allocate a fixed rollout budget of $K{=}16$ into $M{=}4$ entropy-guided trees. Tok is the average number of generated tokens per solution (lower is better). \colorbox{lightgreen}{Green} indicates better results compared with GRPO/GSPO. \colorbox{lightred}{Red} indicates worse results compared with GRPO/GSPO. For accuracy metrics, \textbf{Bold} indicates the highest value in each column
}
\label{tab:main_results}
\end{table*}

\section{Experiments}

\subsection{Experimental Setup}

\paragraph{Models}
We conduct our primary evaluations using three base models from the Qwen3 family~\cite{yang2025qwen3technicalreport}, selected to span different scales and architectures: \texttt{Qwen3-4B-Base/Inst/2507}, \texttt{Qwen3-8B-Base/Inst}, and a Mixture-of-Experts (MoE) model \texttt{Qwen3-30B-A3B-2507}.

To contextualize our results within the state-of-the-art models, we also report performance for a set of strong open-weight reference models, including Qwen3 families (8B, 30B) and the distilled DeepSeek-R1 series (Qwen-based 7B/14B/32B and Llama-based 8B~\cite{deepseekai2025deepseekr1incentivizingreasoningcapability}). Note that these reference models are evaluated "as-is" without further training via \ourmethod.

\paragraph{Benchmarks and Metrics}
Our evaluation spans $10$ benchmarks, grouped into three domains:
(1) \emph{Standard Mathematics}: MATH500~\cite{hendrycks2021measuringmathematicalproblemsolving} serves as the primary baseline.
(2) \emph{Olympiad-Level Competitions}: We use OmniMath~\cite{gao2024omnimathuniversalolympiadlevel} for broad coverage, and a suite of high-difficulty datasets including AIME 2024/2025~\cite{aime2024,aime2025}, HMMT 2025~\cite{balunovic_srimatharena_2025}, BRUMO 2025~\cite{brumo2025}, CMI (CMINMC 2025)~\cite{cmimc2025}, and the OlymMATH-Easy/Hard subsets~\cite{sun2025challengingboundariesreasoningolympiadlevel}.
(3) \emph{Scientific Reasoning}: We use GPQA-Diamond~\cite{rein2023gpqagraduatelevelgoogleproofqa} to assess graduate-level scientific problem-solving.

We report accuracy using Pass@1 for large-scale benchmarks. For smaller, high-variance datasets (at most 40 problems, e.g., AIME, HMMT), we report Pass@32 to ensure statistical stability. Additionally, we measure reasoning efficiency (\textbf{Tok}) using the average number of tokens generated per solution, where a lower count indicates more concise reasoning.

\paragraph{Baselines and Training Details}
We compare against two group-based RL baselines, GRPO and GSPO, and apply \ourmethod on top of both.
Given that standard GRPO often exhibits convergence instability on Mixture-of-Experts (MoE) architectures, we exclusively employ the more robust GSPO baseline for the \texttt{Qwen3-30B-Instruct} MoE model.

In addition to these primary baselines, we further compare \ourmethod with two representative strong alternatives that incorporate step-/tree-structured training signals: DAPO~\cite{yu2025dapoopensourcellmreinforcement} and TreeRL~\cite{hou2025treerlllmreinforcementlearning}.
These methods differ from GRPO/GSPO-style training in their supervision granularity and/or training-time computation.
We therefore report them as complementary baselines and explicitly match training settings where applicable, while also disclosing any deviations in rollout budget or extra scoring required by the method.

To ensure fair comparisons in our primary setting, all GRPO/GSPO and \ourmethod variants are trained on the same $10$k subset sampled from DeepMath103K~\cite{he2025deepmath103klargescalechallengingdecontaminated} under the VERL~\cite{Sheng_2025} framework, with identical optimizer settings, rollout budgets, and decoding configurations.
Complete hyperparameters and implementation details are provided in Appendix~\ref{app:training_details}.

\begin{figure*}[!t]
    \centering
    \includegraphics[width=0.9\linewidth]{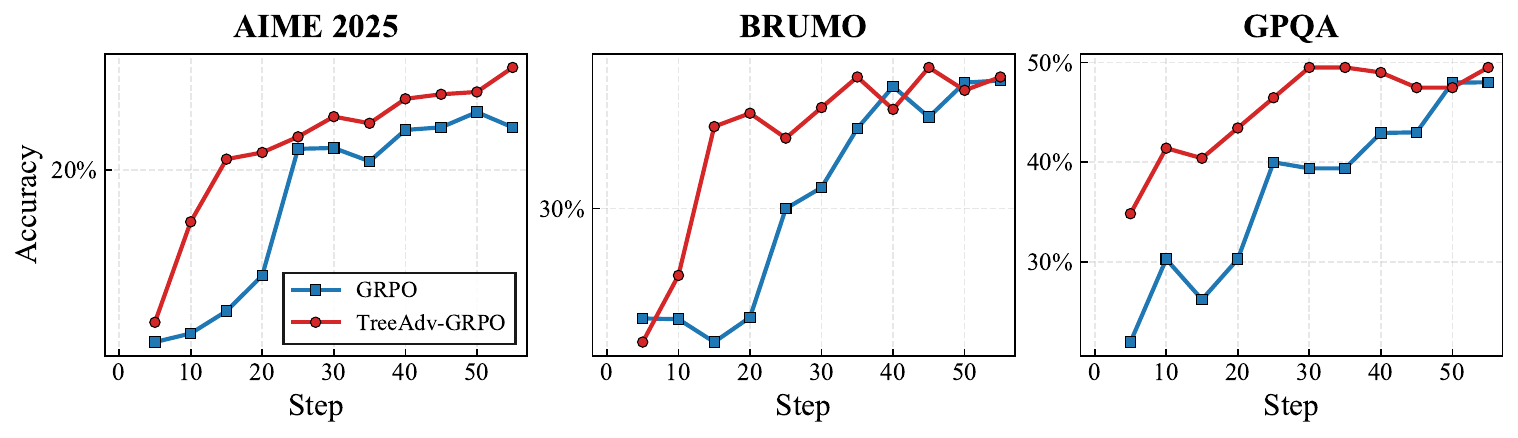}
    \caption{
    Test-set accuracy versus training steps on AIME25, BRUMO, and GPQA.
    All runs are initialized from \textbf{Qwen3-8B-Base} and compare \textbf{GRPO} against \textbf{\ourmethod-GRPO}, illustrating how accuracy evolves during training under the two objectives.
    }
    \label{fig:8B-Base_acc}
\end{figure*}

\begin{figure*}[!t]
    \centering
    \includegraphics[width=0.9\linewidth]{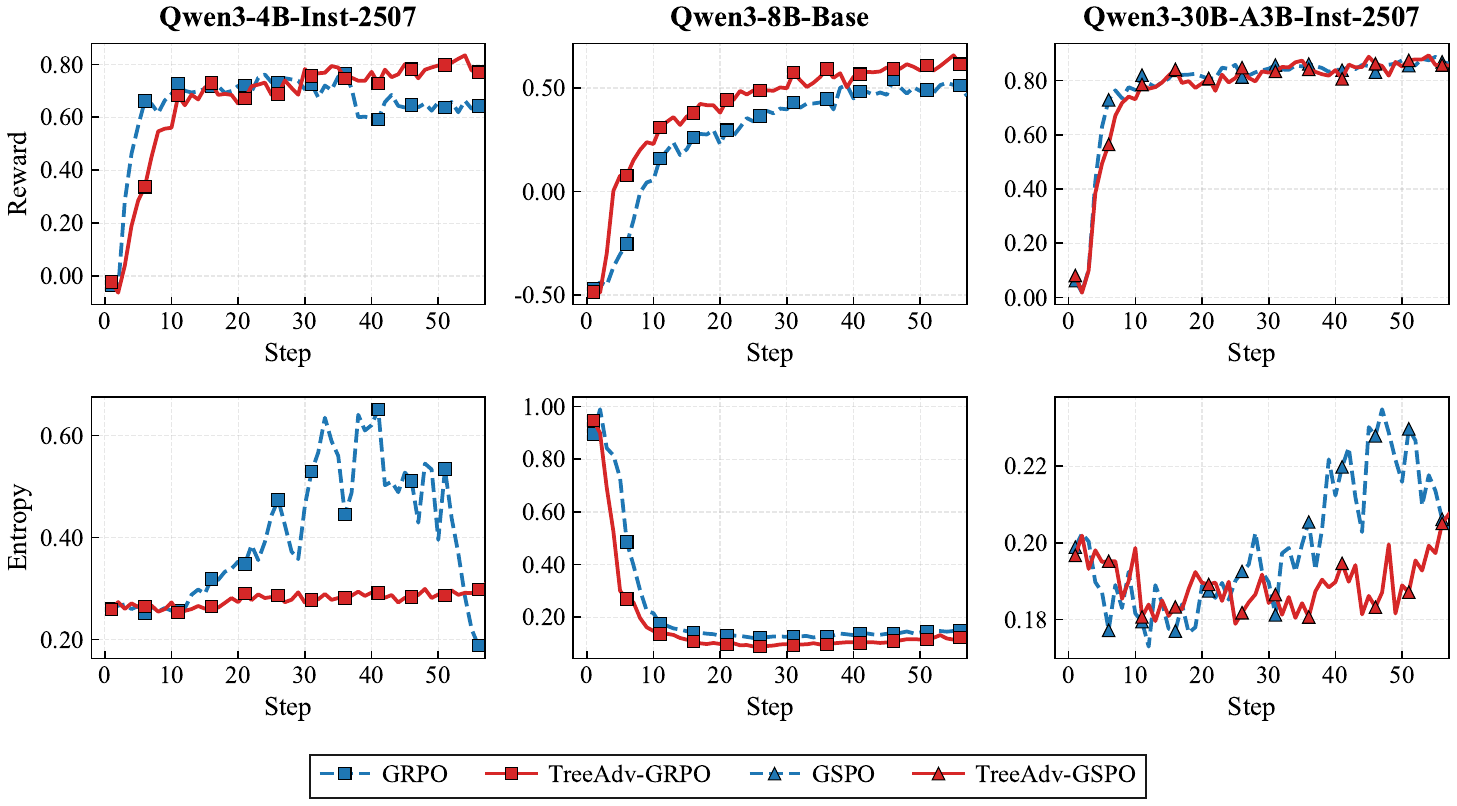}
    \caption{Training and exploration dynamics across model scales. We compare TreeAdv (Ours, red curves) with standard GRPO/GSPO baselines (blue curves) on three model sizes: Qwen3-4B-Instruct-2507, Qwen3-8B-Base, and Qwen3-30B-Instruct-2507. The top row shows reward trajectories, where TreeAdv consistently achieves higher rewards and faster convergence across scales. The bottom row shows entropy trajectories, characterizing how each method maintains exploration while optimizing the policy throughout training.}
    \label{fig:treeadv_reward_entropy_step}
\end{figure*}

\subsection{Main Results}
\label{subsec:main_results}

Table~\ref{tab:main_results} summarizes results on our olympiad-level reasoning suite.
We report average accuracy and \textbf{Tok} (average generated tokens per solution); per-benchmark token statistics are in Appendix~\ref{app:token_analysis}.

Across Blocks~(B)--(C), \ourmethod consistently improves the accuracy--efficiency trade-off.
In most settings, it increases average accuracy while reducing \textbf{Tok}.
For \texttt{Qwen3-8B-Inst Think}, TreeAdv-GRPO improves average accuracy from 60.55\% to 61.99\% with a $\sim$23\% reduction in \textbf{Tok} (15693 $\rightarrow$ 12073), and improves OlymH from 23\% to 27\%.
We note that the gains on some models in Think mode can be modest, likely because the difficulty of DeepMath103K limits additional headroom under terminal-reward supervision.

\begin{table*}[!t]
\centering
\small
\setlength{\tabcolsep}{3.2pt}
\renewcommand{\arraystretch}{1.06}

\newcommand{\mindent}{\hspace{0.6em}}
\newcommand{\sindent}{\hspace{1.2em}}


\resizebox{\textwidth}{!}{%
\begin{tabular}{@{}lcccccccccccc@{}}
\toprule
\multirow{2}{*}{\textbf{Method}}
& \textbf{AIME24} & \textbf{AIME25} & \textbf{BRUMO} & \textbf{CMI}
& \textbf{GPQA} & \textbf{HMMT} & \textbf{MATH}
& \textbf{OlymE} & \textbf{OlymH} & \textbf{Omni}
& \textbf{Avg} $\uparrow$ & \textbf{Tok} $\downarrow$ \\
& {\scriptsize(32p1)} & {\scriptsize(32p1)} & {\scriptsize(32p1)} & {\scriptsize(32p1)}
&  & {\scriptsize(32p1)} & 
&  &  & {\scriptsize}
& {\scriptsize(\%)} & {\scriptsize(\#)} \\
\midrule

\sindent \texttt{Qwen3-8B-Base} & 11.25 & 9.48 & 18.85 & 4.45 & 18.69 & 1.35 & 68.20 & 7.00 & 3.00 & 24.07 & 16.64 & 2667 \\
\sindent \texttt{ w. DAPO} & 31.56 & 25.94 & 38.85 & 16.48 & 41.41 & 14.79 & 87.60 & 18.00 & 4.00 & 35.50 & 31.41 & 6733 \\
\sindent \texttt{ w. TreeRL} & 14.17 & 12.92 & 24.17 & 6.02 & 28.28 & 2.40 & 78.60 & 6.00 & 6.00 & 27.85 & 20.64 & 1945 \\
\sindent \texttt{ w. \ourmetho-GRPO} & 27.71 & 24.79 & 37.19 & 16.09 & 49.49 & 11.77 & 85.60 & 20.00 & 10.00 & 34.03 & \colorbox{lightgreen}{31.67} & 6171 \\

\cmidrule(lr){2-13}
\sindent \texttt{Qwen3-8B-Inst Non-Think} & 28.02 & 21.15 & 30.94 & 14.69 & 44.44 & 10.52 & 83.20 & 23.00 & 4.00 & 35.41 & 29.54 & 3506 \\
\sindent \texttt{ w. DAPO} & 56.25 & 40.00 & 52.92 & 30.31 & 59.60 & 26.98 & 90.80 & 44.00 & 11.00 & 50.52 & 46.24 & 13373 \\
\sindent \texttt{ w. TreeRL}  & 28.33 & 21.67 & 30.42 & 15.63 & 47.47 & 11.88 & 83.80 & 23.00 & 2.00 & 34.62 & 29.88 &  3605 \\
\sindent \texttt{ w. \ourmetho-GRPO} & 60.63 & 49.17 & 55.00 & 33.44 & 54.55 & 29.90 & 92.20 & 58.00 & 12.00 & 48.83 & \colorbox{lightgreen}{49.37} & 9962 \\

\cmidrule(lr){2-13}
\sindent \texttt{Qwen3-4B-Inst-2507} & 61.56 & 45.73 & \textbf{57.81} & 32.03 & 60.10 & 30.31 & 93.40 & 57.00 & 11.00 & 49.55 & 49.85 & 7701 \\
\sindent \texttt{ w. DAPO} & 61.56 & 51.04 & 57.29 & 31.72 & 66.16 & \textbf{31.67} & 93.00 & 58.00 & 10.00 & 50.61 & 51.11 & 6730 \\
\sindent \texttt{ w. TreeRL} & 61.67 & 46.25 & 54.69 & 31.64 & 59.09 & 30.52 & 92.80 & \textbf{59.00} & 17.00 & 48.60 & 50.13 & 7540 \\
\sindent \texttt{ w. \ourmetho-GRPO} & \textbf{61.98} & \textbf{52.81} & 56.98 & \textbf{32.34} & \textbf{64.14} & \textbf{31.67} & \textbf{93.80} & 53.00 & \textbf{18.00} & \textbf{50.72} & \colorbox{lightgreen}{\textbf{51.54}} & 5805 \\
\bottomrule
\end{tabular}
}
\caption{
Results on a suite of reasoning benchmarks comparing additional baselines, including DAPO and TreeRL, against the corresponding TreeAdv-GRPO-trained models. Tok is the average number of generated tokens per solution (lower is better). \colorbox{lightgreen}{Green} indicates better results compared with DAPO/TreeRL, while \colorbox{lightred}{Red} indicates worse results. For accuracy metrics, \textbf{Bold} indicates the highest value in each column.
}
\label{tab:strong_baseline_results}
\end{table*}

Table~\ref{tab:strong_baseline_results} further compares \ourmethod with DAPO and TreeRL under matched supervision and decoding settings.
Notably, DAPO underperforms \ourmethod even with a larger rollout budget (Appendix~\ref{app:training_details}), yielding lower accuracy and higher \textbf{Tok}.
TreeRL exhibits pronounced regime sensitivity: in our evaluation, it provides little to no improvement on base models and mixed Think/Non-Think regimes.
Moreover, even in the instruction-tuned Non-Think setting---commonly considered a favorable regime for TreeRL---it still falls short of \ourmethod, achieving lower accuracy while consuming more generated tokens (\textbf{Tok}).
Overall, \ourmethod delivers more consistent gains across model regimes, improving both accuracy and token efficiency under comparable supervision and decoding settings.

\begin{figure*}[!t]
    \centering
    \includegraphics[width=0.95\linewidth]{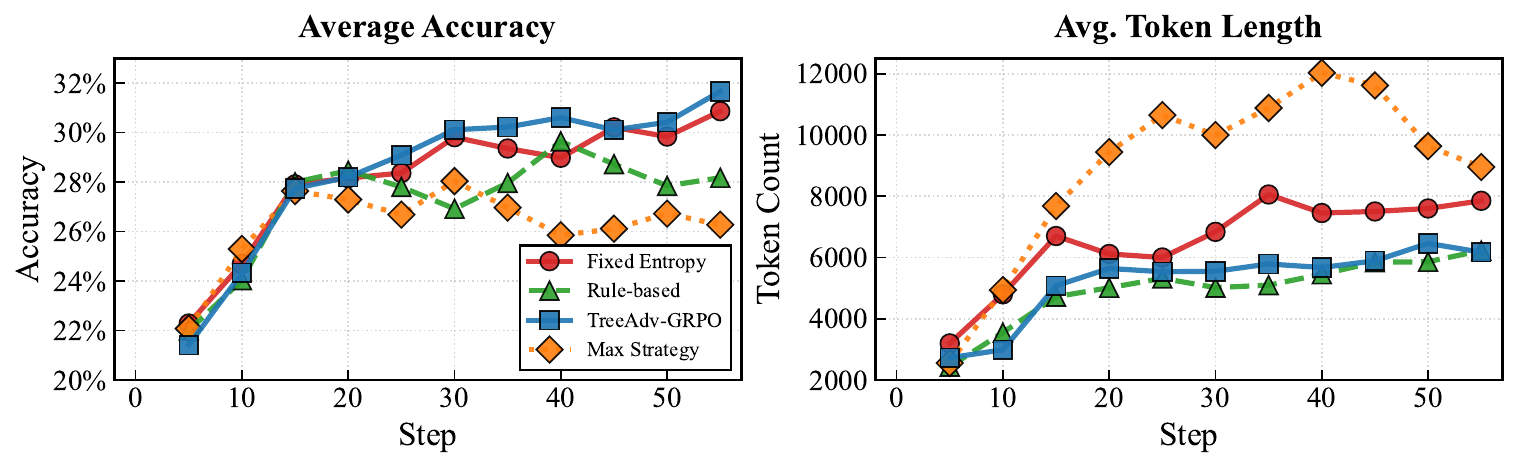}
    \caption{
    Evolution of average accuracy and token length across training steps under four different tree-branching strategies. We compare: Entropy Fixed (fixed entropy threshold $H{=}1.4$), Rule-based (branching at \texttt{.\textbackslash n\textbackslash n} and \texttt{.\textbackslash n} instead of an entropy criterion), TreeAdv-GRPO (ours; default entropy-scheduled branching), and Max Strategy (using max instead of mean when aggregating descendant-leaf advantages). The x-axis denotes the training step; the left y-axis reports the mean accuracy averaged over 10 evaluation benchmarks, while the right y-axis reports the average generated token length.
    }
    \label{fig:abl}
\end{figure*}

\subsection{Training Dynamics and Stability}
\label{subsec:train_dyn}

We next analyze training-time dynamics to understand how the improvements emerge.
Figure~\ref{fig:treeadv_reward_entropy_step} compares \ourmethod with GRPO/GSPO across model scales, reporting reward (top) and entropy (bottom) over training steps; entropy serves as a proxy for policy sharpness and training-time volatility.

On \texttt{Qwen3-8B-Base}, \ourmethod achieves faster early reward gains and a quicker entropy drop, after which the reward gap narrows later in training, consistent with earlier policy concentration under the same budget.
On instruction-tuned models, \ourmethod often shows a short delay in early reward growth but consistently catches up and surpasses GRPO/GSPO, with typically smoother entropy trajectories, suggesting lower-variance updates from token-/segment-level advantage redistribution.

We further conduct checkpoint analyses on \texttt{Qwen3-8B-Base}.
Figure~\ref{fig:8B-Base_acc_all} reports Pass@1 accuracy over training steps across all benchmarks, and Figure~\ref{fig:8B-Base_acc} zooms in on AIME~2025, BRUMO, and GPQA using dense checkpoints saved every 5 steps; accuracy broadly tracks reward, where \ourmethod improves steadily while GRPO shows a mid-training regression.
Finally, Figure~\ref{fig:8B-Base_acc_token} jointly plots Pass@1 accuracy (solid) and average output length (dashed); between steps 20 and 40, \ourmethod improves accuracy without increasing output length, indicating gains beyond simply generating longer solutions.

\section{Ablation}
\label{sec:ablation}

We ablate three components of \ourmethod under identical training data, rollout budget, decoding, and optimization settings (Figure~\ref{fig:abl}).
Specifically: (i) \textbf{Rule-based branching}, which replaces entropy-guided branching with delimiter-based splits at tokens such as \texttt{.\textbackslash n\textbackslash n} and \texttt{.\textbackslash n}; (ii) \textbf{Max Strategy}, which replaces the mean in Eq.~\ref{eq:token_adv} with a max operator when aggregating descendant leaf advantages; and (iii) \textbf{Entropy Fixed}, which fixes the branching threshold at $H{=}1.4$ throughout training instead of using our schedule.

\begin{figure*}[!t]
    \centering
    \includegraphics[width=0.95\linewidth]{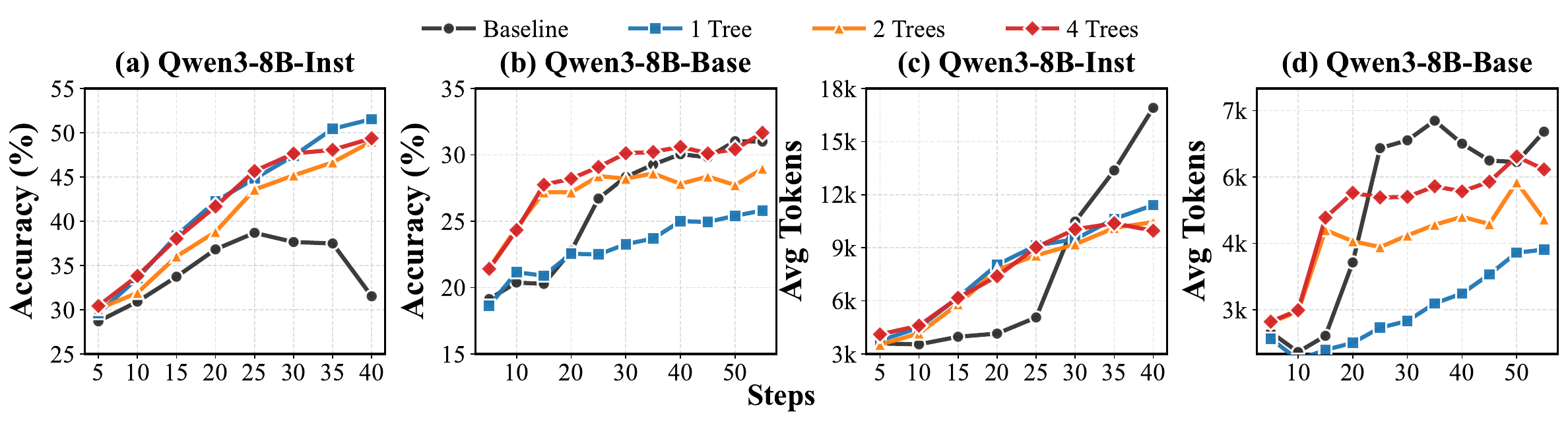}
    \caption{
     Performance and cost analysis across training steps. (a) and (b) illustrates the Average Accuracy trends, while (c) and (d) shows the corresponding Average Token consumption. The left column depicts the Qwen3-8B-Inst Non-think model, and the right column depicts the Qwen3-8B-Base model. Curves represent comparisons between the Baseline and varying tree search strategies (1 Tree, 2 Trees, and 4 Trees)
    }
    \label{fig:ntree}
\end{figure*}

Figure~\ref{fig:abl} yields three takeaways.
First, \emph{where} we branch matters: rule-based branching shows limited late-stage accuracy gains, suggesting that formatting boundaries are weaker proxies for high-uncertainty decision points than entropy-guided splits.
Second, \emph{how} we aggregate leaf advantages matters: Max Strategy improves faster early on but plateaus, indicating that emphasizing the single best descendant can produce a less stable training signal than averaging.
Third, the entropy schedule has a stronger impact on efficiency than on accuracy: compared with Entropy Fixed, our adaptive threshold yields lower token usage in later training while maintaining comparable accuracy.

\section{Analysis}
\label{sec:analysis}

\subsection{Allocating a fixed rollout budget: sharing vs.\ exploration}
Figure~\ref{fig:ntree} analyzes training dynamics under a fixed rollout budget of $K{=}16$ completed trajectories per question.
GRPO/GSPO sample $K$ independent rollouts without shared prefixes.
Our tree-structured variants allocate the same budget into $M \in \{1,2,4\}$ entropy-guided trees: smaller $M$ increases prefix sharing, while larger $M$ increases root-level diversity.

The optimal allocation is regime-dependent.
On the instruct model (Figure~\ref{fig:ntree}a), $M{=}1$ yields the best and most stable accuracy gains, with $M{=}2$ and $M{=}4$ slightly behind.
On the base model (Figure~\ref{fig:ntree}b), $M{=}4$ achieves the best final accuracy, while $M{=}1$ plateaus early.
Overall, base models benefit more from root-level exploration, whereas instruct models benefit more from lower-variance learning signals enabled by stronger prefix sharing.

\subsection{Accuracy versus token usage and stability}
Token-usage trends indicate that gains are not simply driven by longer generations.
On the instruct setting, GRPO becomes increasingly verbose at later steps while accuracy decreases (Figures~\ref{fig:ntree}a,c), whereas tree-structured variants keep token usage more controlled and are less prone to late-stage regression.
On the base setting, token usage increases with larger $M$ (Figure~\ref{fig:ntree}d), reflecting the cost of broader exploration; notably, the best-performing configuration ($M{=}4$) is also the most token-expensive.
Appendix~\ref{app:rollout-token-stats} (Figure~\ref{fig:token_rollout_count}) reports rollout token statistics across additional model scales.

\section{Conclusion}
\label{sec:conclusion}

This work studies how to improve RL fine-tuning under a fixed rollout budget by changing how rollouts are sampled and how advantage is assigned.
We propose \ourmethod, including tree-structured sampling which organizes $K$ completed trajectories into $M$ entropy-guided trees, and a structure-aware token-level advantage redistribution that uses outcome differences among continuations sharing the same prefix.
This yields more comparable learning signals on shared segments and reduces spurious reinforcement of non-informative tokens.

Across model families, our method consistently increases average accuracy while \emph{reducing} average token usage under the same rollout budget, demonstrating that gains are not driven by longer generations.
Training is also more stable across epochs and settings, with lower variance and fewer late-stage regressions.

\section*{Limitations}

Our experiments primarily focus on math-heavy long-form reasoning benchmarks and a small set of Qwen-family backbones. While \ourmethod is not inherently math-specific, additional evaluations on other domains (e.g., code, tool use, dialogue) and model families are needed to characterize generality.

\ourmethod also depends on practical rollout-tree construction choices (entropy thresholding, branching constraints, and a no-branch token list).
These heuristics may require tuning across decoding settings, and our entropy estimate is approximated from top-$k$ probabilities.

Finally, \ourmethod introduces extra systems complexity (tree/forest rollout management and segment-level aggregation). Although it often reduces redundant generation, realizing wall-clock gains may require careful implementation and scaling studies.

\bibliography{custom}
\appendix
\newpage
\section{Appendix}


\subsection{Entropy Details}
\label{app:entropy}

\paragraph{No-branch token list.}
In practice, some high-entropy positions correspond primarily to formatting or structural tokens (e.g., whitespace, LaTeX control symbols, or bracket/brace delimiters). Branching at such positions rarely yields semantically meaningful alternatives and can unnecessarily increase the branching factor. We therefore maintain a \emph{no-branch token list}: even when the entropy criterion $H(s_t)>\tau$ is satisfied, we suppress branching if the token at position $t$ matches any entry in this list (see the tcolorbox below).

\begin{tcolorbox}[breakable]
\textbackslash, \$, \textbackslash n, \textbackslash r, \textvisiblespace, \_, \textvisiblespace\textvisiblespace, :, \textbackslash(, \textbackslash), \textbackslash[, \textbackslash], \textbackslash\{,  \textbackslash\}, (, ), [, ], \{, \}
\end{tcolorbox}

\paragraph{Entropy estimation with truncated log-probabilities.}
We obtain rollouts and token log-probabilities via vLLM, which returns log-probabilities only for the top-$k$ candidates (with $k=20$ in our setup). As a result, we approximate the token entropy using this truncated support rather than the full vocabulary:
\begin{equation}
\tilde{H}(s_t)
\;=\;
-\sum_{a \in \mathcal{A}_{20}(s_t)}
\pi_{\text{old}}(a \mid s_{<t})
\log \pi_{\text{old}}(a \mid s_{<t}),
\end{equation}
where $\mathcal{A}_{20}(s_t)$ denotes the set of top-20 tokens returned by vLLM at state $s_t$ (tokens outside $\mathcal{A}_{20}(s_t)$ are treated as having zero probability under this approximation). We use $\tilde{H}(s_t)$ for entropy-guided branching in all experiments.

\subsection{Training Details}
\label{app:training_details}

\paragraph{Training data.}
All RL experiments are trained on the same supervision pool to enable a controlled comparison. We sample a fixed 10K subset from the DeepMath103K corpus and reuse it for all GRPO, GSPO, and \ourmethod runs. The subset is \emph{not} reshuffled across methods: each method observes the same examples in the same order. This shared data schedule simplifies the analysis of optimization dynamics and helps isolate the effect of the learning algorithm.

\paragraph{Optimization and learning-rate schedule.}
All methods are implemented in the VERL framework, with optimization settings matched across objectives whenever possible. For all \texttt{Base} models, we use a peak learning rate of $5\times10^{-6}$; for all \texttt{Inst} models, we use $1\times10^{-7}$. The minimum learning rate is set to $1\times10^{-7}$ in all runs. Training lasts 57 optimization steps with a global batch size of 512, corresponding to approximately 3 epochs over the 10K subset. We use 2 warmup steps with a linear ramp from 0 to the peak learning rate, followed by cosine decay to the minimum learning rate. This schedule (total steps, warmup length, and cosine decay) is kept identical for GRPO, GSPO, \ourmethod-GSPO and \ourmethod-GRPO.

\paragraph{Checkpointing and model selection.}
We save checkpoints every 5 optimization steps throughout training. For each method and model, we evaluate every saved checkpoint and report the best-performing checkpoint on the held-out benchmarks in the main results. This early-stopping-style selection is applied uniformly to GRPO, GSPO, \ourmethod-GSPO and \ourmethod-GRPO, ensuring that each method is compared at its best point along an otherwise shared training trajectory.

\paragraph{\ourmethod-specific hyperparameters.}
\ourmethod introduces a small set of additional hyperparameters beyond the GRPO/GSPO baselines. We use an initial entropy threshold of $H_0 = 1.4$ to determine branching points in the tree. The threshold is annealed with a decrement of $0.05$ over training, with a floor at $H_{\min} = 1.0$. Unless otherwise specified, for each prompt \ourmethod constructs 4 trees. Each prompt is allotted a total of 16 rollouts, distributed across these trees, and this tree/rollout configuration is held fixed across all \ourmethod experiments. To avoid premature branching at the very beginning of generation, we additionally impose an earliest-branching constraint: branching is only allowed at the first position that satisfies the entropy threshold \emph{after} encountering one of the following delimiters in the generated text: \texttt{.\textbackslash n\textbackslash n}, \texttt{,~}, or \texttt{.\textbackslash n}.

Our entropy-guided branching is inspired by prior work on uncertainty-guided exploration~\cite{cao2026bangbuckprocessreward}. 
That work further reports an empirical analysis of token-level entropy distributions across datasets and difficulty strata (see its appendix A.12), which motivates our use of an annealed entropy threshold and the earliest-branching constraint.

\paragraph{Sampling for rollouts and inference.}
During training rollouts, we use top-$k$ and nucleus (top-$p$) sampling with $k=20$ and $p=0.7$. We use the same sampling configuration for test-time inference on the evaluation benchmarks (top-$k$ = $20$, top-$p$ = $0.7$) to keep generation stochasticity consistent between training and evaluation.

\paragraph{Reward design and other settings.}
Reward design and other RL-specific settings (e.g., correctness scoring, shaping, and penalty terms) follow the recommended configuration in the DeepMath103K release. We use the same reward function for GRPO, GSPO, \ourmethod-GSPO and \ourmethod-GRPO so that observed differences are attributable to the training objective and advantage construction rather than changes in supervision. To prevent excessively long generations, we apply a length penalty consistently across all experiments: if the output exceeds 16K tokens, we set the original reward to $-1$.

\paragraph{DAPO Training}
For fair comparison, we follow DAPO's recommended training recipe and keep all optimization hyperparameters unchanged; the only difference is that we use a $2\times$ larger rollout budget to instantiate the tree-structured rollouts (all remaining settings strictly match those in DAPO).

\paragraph{TreeRL}
For fair comparison, we adopt TreeRL's default recommended configuration $(M,N,L,T)=(6,2,1,2)$; aside from this, we keep the training pipeline and all remaining hyperparameters exactly the same as in the above setup.

\paragraph{Computational resources.}
The comprehensive experimental suite, including all training phases, hyperparameter tuning, and inference evaluations, was executed on a high-performance cluster equipped with \textbf{128 NVIDIA A800 GPUs}. The total computational cost for this study, covering all baselines and our proposed method, amounted to approximately \textbf{1,500 GPU hours}.

\subsection{Test datasets details}
\label{app:test_dataset_details}
We evaluate our models on a diverse collection of ten benchmarks, ranging from standard high school mathematics to graduate-level scientific reasoning and recent Olympiad competitions. The details of each dataset are as follows:

\paragraph{Standard Mathematics}
\begin{itemize}
    \item \textbf{MATH500}: A widely adopted subset of the MATH dataset, consisting of 500 randomly sampled problems. It covers seven subject areas (e.g., algebra, number theory, calculus) and serves as a baseline for measuring standard mathematical problem-solving capabilities.
\end{itemize}

\paragraph{Olympiad-Level Competitions}
This category includes both broad-coverage benchmarks and specific high-difficulty contests from the 2024--2025 season to prevent data contamination and test frontier reasoning.
\begin{itemize}
    \item \textbf{OmniMath}: A comprehensive Olympiad-level benchmark designed to assess advanced mathematical reasoning. We utilize the full test set consisting of 4,428 problems, which aggregates diverse competitions to ensure broad coverage of Olympiad-style challenges.
    
    \item \textbf{AIME 2024 \& 2025}: The American Invitational Mathematics Examination (AIME) is a high-prestige intermediate Olympiad. We include all problems from both the AIME I and AIME II exams for each year, resulting in 30 problems for AIME 2024 and 30 problems for AIME 2025. These questions require integer answers between 000 and 999.
    
    \item \textbf{HMMT 2025}: The Harvard-MIT Mathematics Tournament (February 2025). We evaluate on a subset of 30 problems, specifically comprising the Algebra \& Number Theory, Geometry, and Combinatorics subject rounds (10 problems each). This dataset represents one of the most challenging undergraduate-organized competitions for high school students.
    
    \item \textbf{BRUMO 2025}: The Brown University Math Olympiad (2025). We evaluate on a total of 30 problems, combining the complete Individual Round (15 problems) and the Team Round (15 problems). This dataset serves as a fresh evaluation benchmark with diverse problems across algebra, combinatorics, geometry, and number theory.
    
    \item \textbf{CMINMC 2025}: The Carnegie Mellon Informatics and Mathematics Competition (2025). We evaluate on a total of 40 problems, comprising the three Individual Rounds (Algebra \& Number Theory, Combinatorics \& Computer Science, and Geometry) and the Team Round, with 10 problems each. This dataset covers a hybrid of mathematical and algorithmic reasoning tasks.
    
    \item \textbf{OlymMATH}: A benchmark designed to challenge the reasoning boundaries of LLMs. We evaluate on two sampled subsets to assess performance across difficulty gradients:
    \begin{itemize}
        \item \textbf{OlymMATH-Easy}: A subset of \textbf{100} problems representing entry-level Olympiad difficulty.
        \item \textbf{OlymMATH-Hard}: A subset of \textbf{100} high-complexity problems, selected to test the models' capability in handling intricate logical deductions and multi-step reasoning.
    \end{itemize}
\end{itemize}

\paragraph{Scientific Reasoning}
\begin{itemize}
    \item \textbf{GPQA-Diamond}: A subset of the Graduate-Level Google-Proof Q\&A benchmark~\cite{rein2023gpqagraduatelevelgoogleproofqa}. It contains 198 multiple-choice questions written by domain experts in biology, physics, and chemistry. This dataset is designed to be difficult even for human experts to answer without access to external resources, serving as a proxy for high-level scientific reasoning.
\end{itemize}

\subsection{Test Datasets Token Statistics}
\label{app:token_analysis}

This subsection provides detailed token-usage statistics corresponding to Table~\ref{tab:main_results}.
Table~\ref{tab:token_efficiency} reports, for each benchmark, the average number of generated tokens per solution (\#; lower is better), which serves as a fine-grained proxy for generation-time compute and verbosity.
We organize results into three groups: \textbf{(A)} representative SOTA open-weight models to calibrate typical token budgets; \textbf{(B)} controlled comparisons on base models; and \textbf{(C)} controlled comparisons on instruct models.
Within each controlled comparison group, bold numbers indicate the more token-efficient method (i.e., fewer generated tokens).
These per-benchmark statistics complement the aggregate \textbf{Tok} reported in the main table and help verify that improvements in the accuracy--efficiency frontier reflect broadly reduced redundant generation rather than being driven by a small subset of tasks.

\begin{table*}[!t]
\centering
\small
\setlength{\tabcolsep}{3.5pt}
\renewcommand{\arraystretch}{1.06}

\newcommand{\mindent}{\hspace{0.6em}}
\newcommand{\sindent}{\hspace{1.2em}}

\resizebox{\textwidth}{!}{%
\begin{tabular}{@{}lccccccccccc@{}}
\toprule
\multirow{2}{*}{\textbf{Model / Method}}
& \textbf{AIME24} & \textbf{AIME25} & \textbf{BRUMO} & \textbf{CMI}
& \textbf{GPQA} & \textbf{HMMT} & \textbf{MATH}
& \textbf{OlymE} & \textbf{OlymH} & \textbf{Omni}
& \textbf{Avg} \\
& {\scriptsize(\#)} & {\scriptsize(\#)} & {\scriptsize(\#)} & {\scriptsize(\#)}
& {\scriptsize(\#)} & {\scriptsize(\#)} & {\scriptsize(\#)}
& {\scriptsize(\#)} & {\scriptsize(\#)} & {\scriptsize(\#)}
& {\scriptsize(\#)} \\
\midrule

\multicolumn{12}{@{}l@{}}{\textit{\textbf{(A) SOTA open models (Qwen \& DeepSeek)}}} \\
\addlinespace[0.15em]
\mindent \texttt{Qwen3-32B-Inst Think} & 13137 & 15680 & 13993 & 18622 & 8086 & 17736 & 4485 & 15424 & 20951 & 12800 & 14091 \\
\mindent \texttt{Qwen2.5-7B-Inst} & 1999 & 1344 & 1251 & 1190 & 724 & 1151 & 661 & 1096 & 1252 & 1239 & 1191 \\
\mindent \texttt{Qwen2.5-Math-7B-Inst} & 1464 & 1290 & 1243 & 1284 & 1102 & 1216 & 662 & 1401 & 1260 & 1042 & 1196 \\
\mindent \texttt{Qwen2.5-QwQ-32B} & 13902 & 15970 & 13799 & 18656 & 8872 & 17481 & 4190 & 15941 & 20613 & 12331 & 14176 \\
\mindent \texttt{DeepSeek-R1-Distill-Qwen-7B} & 12629 & 13648 & 11558 & 15544 & 6922 & 15275 & 3370 & 15188 & 17378 & 10708 & 12222 \\
\mindent \texttt{DeepSeek-R1-Distill-Qwen-14B} & 11190 & 13551 & 11242 & 15774 & 6965 & 15552 & 3812 & 14653 & 17568 & 10873 & 12118 \\
\mindent \texttt{DeepSeek-R1-Distill-Qwen-32B} & 11339 & 13351 & 10265 & 14892 & 6459 & 14968 & 3715 & 12929 & 17524 & 10586 & 11603 \\
\mindent \texttt{DeepSeek-R1-Distill-Llama-8B} & 13772 & 14188 & 12767 & 16124 & 7727 & 15472 & 3570 & 15713 & 18025 & 11176 & 12854 \\

\cmidrule(lr){1-12}
\multicolumn{12}{@{}l@{}}{\textit{\textbf{(B) Our \ourmetho on base models}}} \\
\addlinespace[0.10em]
\sindent \texttt{Qwen3-8B-Base} & 4336 & 2948 & 3264 & 2753 & 2287 & 3054 & 1163 & 1912 & 2416 & 2540 & 2667 \\ 
\sindent \texttt{ w. GRPO} & 9466 & 8919 & 6909 & 8600 & 1917 & 8807 & 2004 & 9182 & 8533 & 5911 & 7025 \\
\sindent \texttt{ w. \ourmetho-GRPO} & 8256 & 7218 & 6460 & 7057 & 3909 & 7385 & 2047 & 7321 & 6818 & 5244 & \colorbox{lightgreen}{6171} \\
\sindent \texttt{ w. GSPO} & 13527 & 11894 & 10158 & 11924 & 1951 & 13309 & 2202 & 11377 & 11868 & 7615 & 9583 \\
\sindent \texttt{ w. \ourmetho-GSPO} & 12194 & 9695 & 8732 & 9000 & 3818 & 11381 & 2345 & 12064 & 7872 & 6086 & 8319 \\
\cmidrule(lr){2-12}
\sindent \texttt{Qwen3-4B-Base} & 4343 & 3217 & 2481 & 2567 & 2096 & 3348 & 1377 & 3003 & 1067 & 2780 & 2628 \\
\sindent \texttt{ w. GRPO} & 5963 & 3141 & 3792 & 2531 & 1760 & 3386 & 1001 & 2015 & 1997 & 2360 & 2794 \\
\sindent \texttt{ w. \ourmetho-GRPO} & 8222 & 6941 & 6357 & 6187 & 1653 & 7920 & 2083 & 7398 & 6962 & 5220 & 5894 \\
\sindent \texttt{ w. GSPO} & 12259 & 8626 & 6509 & 6932 & 3138 & 7693 & 1742 & 7535 & 4425 & 4597 & 6346 \\
\sindent \texttt{ w. \ourmetho-GSPO} & 14467 & 10840 & 9129 & 8947 & 3222 & 11976 & 2906 & 10055 & 8232 & 6285 & 8606 \\

\cmidrule(lr){1-12}
\multicolumn{12}{@{}l@{}}{\textit{\textbf{(C) Our \ourmethod on instruct models}}} \\

\sindent \texttt{Qwen3-4B-Inst Think} & 14441 & 17233 & 15469 & 20303 & 9211 & 18488 & 5029 & 17475 & 21994 & 13578 & 15322 \\
\sindent \texttt{ w. GRPO} & 14382 & 17331 & 15009 & 19634 & 9184 & 18190 & 5020 & 16400 & 21603 & 13186 & 15000 \\
\sindent \texttt{ w. \ourmetho-GRPO} & 11927 & 12930 & 15474 & 13243 & 17687 & 8008 & 16230 & 4552 & 15313 & 19030 & 13439\\

\cmidrule(lr){2-12}
\sindent \texttt{Qwen3-4B-Inst Non-Think} & 5381 & 4040 & 3170 & 4163 & 2622 & 3476 & 1108 & 3462 & 3002 & 2899 & 3333 \\
\sindent \texttt{ w. GRPO} & 12612 & 12473 & 7313 & 13816 & 4959 & 12984 & 1958 & 11139 & 11616 & 8225 & 9710 \\
\sindent \texttt{ w. \ourmetho-GRPO} & 11072 & 9878 & 6168 & 11215 & 4575 & 10783 & 1685 & 8787 & 9533 & 6459 & \colorbox{lightgreen}{8015} \\

\cmidrule(lr){2-12}
\sindent \texttt{Qwen3-8B-Inst Think} & 15001 & 18017 & 16378 & 20775 & 9862 & 20066 & 5472 & 18667 & 23282 & 14481 & 16200 \\
\sindent \texttt{ w. GRPO} & 15039 & 17524 & 16066 & 20363 & 9450 & 19604 & 5317 & 17035 & 22212 & 14319 & 15693 \\
\sindent \texttt{ w. \ourmetho-GRPO} & 11825 & 13736 & 12112 & 15503 & 6978 & 14977 & 4093 & 13615 & 17324 & 10570 & \colorbox{lightgreen}{12073} \\

\cmidrule(lr){2-12}
\sindent \texttt{Qwen3-8B-Inst Non-Think} & 5973 & 4078 & 3369 & 4149 & 2530 & 4190 & 1224 & 3553 & 2912 & 3081 & 3506 \\
\sindent \texttt{ w. GRPO} & 13540 & 13816 & 12089 & 17871 & 12874 & 16933 & 3162 & 15457 & 17402 & 10585 & 13373 \\
\sindent \texttt{ w. \ourmetho-GRPO} & 12361 & 12991 & 10416 & 15377 & 5580 & 15064 & 2132 & 13766 & 17114 & 9291 & \colorbox{lightgreen}{11409} \\
\sindent \texttt{ w. GSPO} & 11021 & 11076 & 8209 &	13229 & 4303 & 11629 & 2044 & 11864 & 13011 & 7814 & 9420\\
\sindent \texttt{ w. \ourmetho-GSPO} & 12893 & 13378 & 10657 & 15356 & 5469 & 14962 & 2316 & 15373 & 17553 & 9543 & 11750 \\

\cmidrule(lr){2-12}
\sindent \texttt{Qwen3-4B-Inst-2507} & 8814 & 8379 & 7095 & 9820 & 5374 & 9555 & 1726 & 9893 & 10051 & 6304 & 7701 \\
\sindent \texttt{ w. GRPO}  & 11093 & 11055 & 8882 & 14011 & 6481 & 15091 & 2023 & 12199 & 15286 & 7852 & 10397 \\
\sindent \texttt{ w. \ourmetho-GRPO} & 6072 & 6346 & 5483 & 6847 & 4621 & 6810 & 1806 & 7423 & 7371 & 5268 & \colorbox{lightgreen}{5805} \\
\sindent \texttt{ w. GSPO}  & 7129 & 7014 & 5989 & 7665 & 4435 & 8188 & 1605 & 7925 & 7975 & 5396 &  \colorbox{lightgreen}{6332}\\
\sindent \texttt{ w. \ourmetho-GSPO} & 6265 & 6960 & 5820 & 7470 & 4978 & 7262 & 1886 & 7725 & 7936 & 5670 &  6197\\

\cmidrule(lr){2-12}
\sindent \texttt{Qwen3-30B-A3B-2507} & 6114 & 6895 & 6031 & 7666 & 5142 & 7638 & 1610 & 6890 & 9212 & 5713 & 6291 \\
\sindent \texttt{ w. GSPO} & 5712 & 6185 & 5502 & 6807 & 4314 & 6840 & 1414 & 6499 & 7930 & 5125 & 5633 \\
\sindent \texttt{ w. \ourmetho-GSPO} & 5641 & 6390 & 5601 & 7039 & 4403 & 6879 & 1450 & 6388 & 8532 & 5156 & \colorbox{lightgreen}{5748} \\
\bottomrule
\end{tabular}
}
\caption{
Token efficiency on olympiad-level reasoning benchmarks. We report the average number of generated tokens per solution (lower is better).
\colorbox{lightgreen}{Green} indicates the more efficient method (lower token count) within each controlled comparison group.
}
\label{tab:token_efficiency}
\end{table*}

\subsection{Rollout Token Statistics}
\label{app:rollout-token-stats}

\begin{figure*}[!t]
    \centering
    \includegraphics[width=0.99\linewidth]{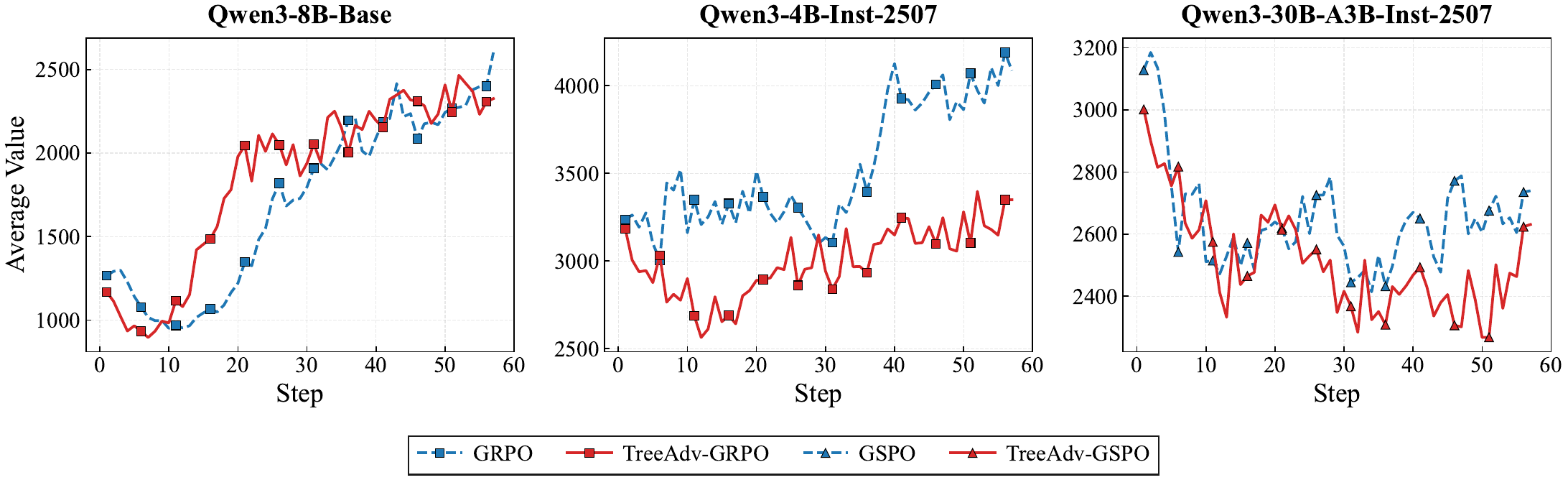}
    \caption{
     Comparative analysis of TreeAdv versus GRPO and GSPO across model scales. The curves report the \emph{average number of tokens per question during training rollouts} (averaged over the fixed rollout budget per question) for Qwen3-4B-Inst-2507, Qwen3-8B-Base, and Qwen3-30B-Inst-2507. TreeAdv (red) yields token usage that is comparable to or lower than the baselines (blue), indicating that its gains are not driven by longer rollouts.
    }
    \label{fig:token_rollout_count}
\end{figure*}

Figure~\ref{fig:token_rollout_count} shows the training-time evolution of the \emph{average number of tokens per question} measured over the rollouts collected during training. 
On Qwen3-4B-Inst-2507, the GRPO baseline exhibits a clear trend toward higher token usage as training progresses, while TreeAdv keeps the per-question token count comparatively stable. 
On Qwen3-8B-Base and Qwen3-30B-Inst-2507, TreeAdv follows token-usage dynamics similar to the corresponding baselines (GRPO and GSPO, respectively). 
Overall, the results indicate that TreeAdv does not rely on systematically longer generations during training, and its behavior is consistent across model scales. 
Since the training-time compute cost is approximately proportional to the number of processed tokens (ignoring first-token latency and other constant overheads), these token-usage trends also suggest comparable training efficiency between TreeAdv and the baselines.

\subsection{Training Accuracy and Output Length on Qwen3-8B-Base}

\begin{figure}[!t]
    \centering
    \includegraphics[width=0.95\linewidth]{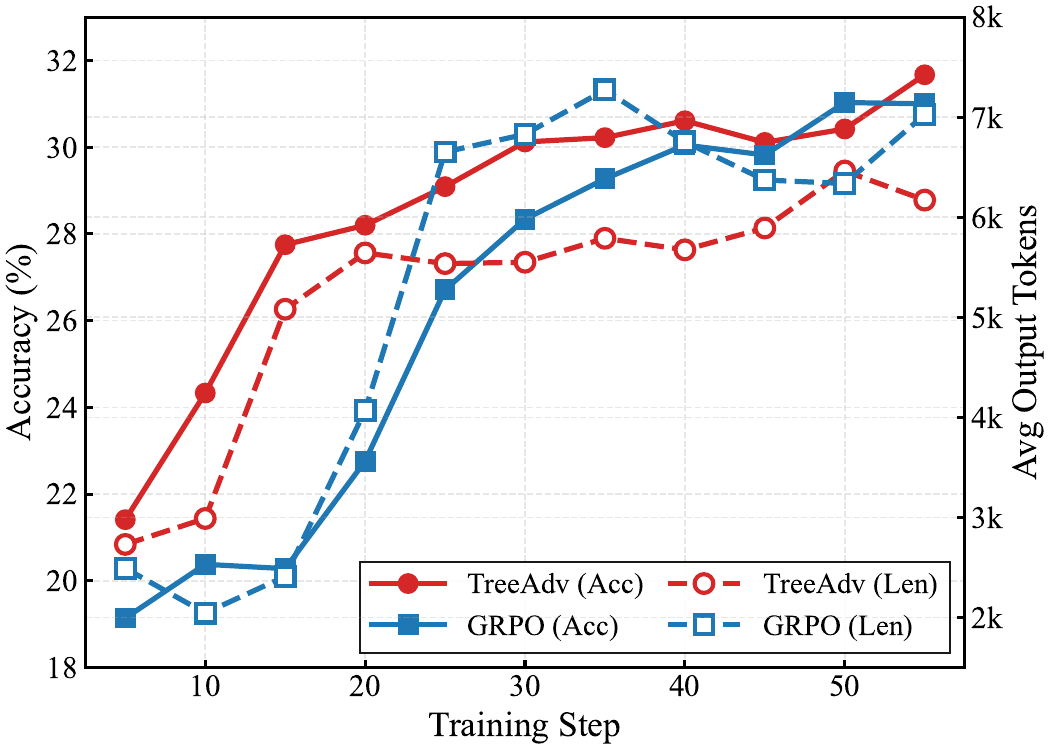}
    \caption{
    Training dynamics comparison between TreeAdv (Ours) and GRPO (Baseline). The solid lines represent the accuracy (left axis), while the dashed lines indicate the average output token length (right axis). TreeAdv (red) consistently outperforms the baseline (blue) in accuracy while maintaining comparable reasoning lengths.
    }
    \label{fig:8B-Base_acc_token}
\end{figure}

We analyze the training dynamics of TreeAdv (ours) and GRPO (baseline) on Qwen3-8B-Base from two complementary views.
Figure~\ref{fig:8B-Base_acc_token} presents an aggregate comparison: solid curves show Pass@1 accuracy over training steps (left axis), while dashed curves report the average generated output length in tokens (right axis).
TreeAdv achieves higher accuracy than GRPO throughout training and improves more steadily.
Meanwhile, the output-length curves remain close, suggesting that the gains are not primarily explained by systematically longer generations, but by more effective optimization under comparable output lengths.

\begin{figure*}[!t]
    \centering
    \includegraphics[width=0.95\linewidth]{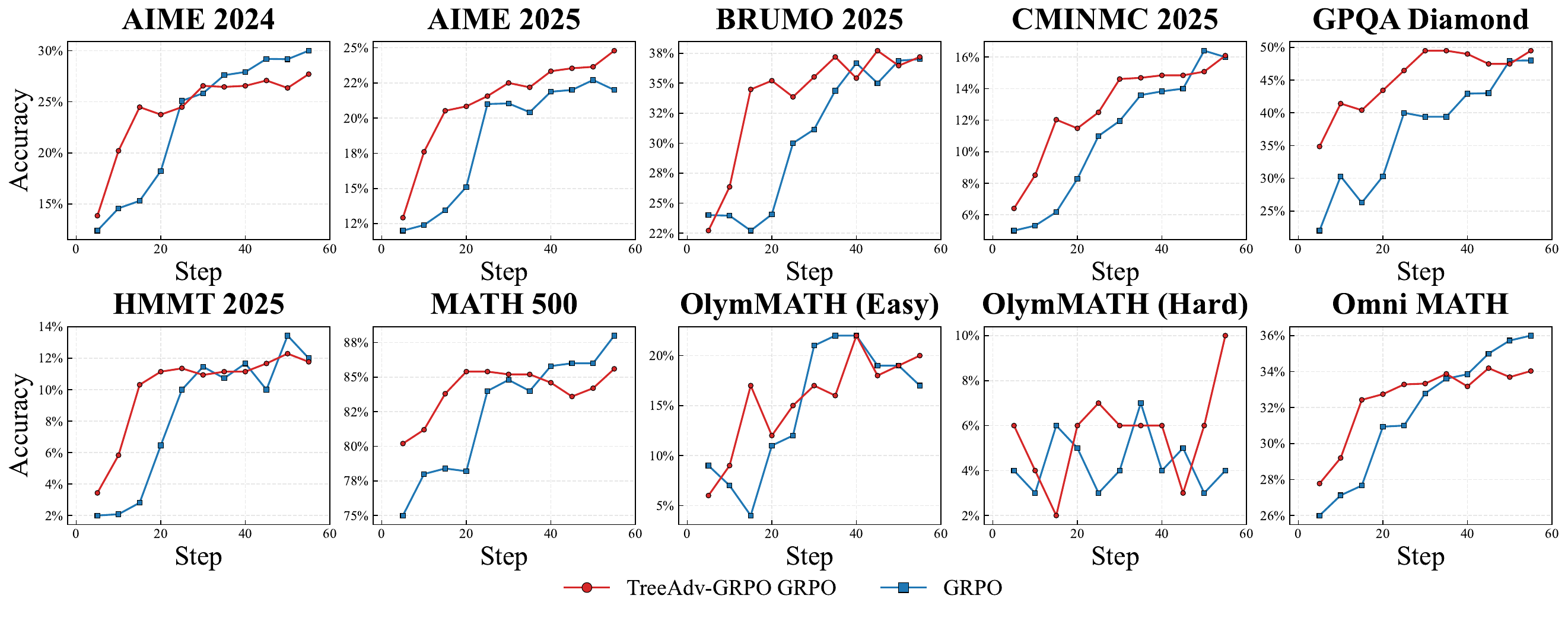}
    \caption{
    Training dynamics on ten mathematical reasoning benchmarks. We compare the performance of our proposed TreeAdv-GRPO (red circles) against the GRPO baseline (blue squares). The curves illustrate the Pass@1 accuracy evolution over training steps. Across diverse datasets—including challenging competitions like AIME 2024/2025 and comprehensive benchmarks like Omni MATH and GPQA Diamond—TreeAdv-GRPO consistently demonstrates superior convergence speed and higher final accuracy compared to the baseline.
    }
    \label{fig:8B-Base_acc_all}
\end{figure*}

To verify that this advantage is not driven by a single dataset, Figure~\ref{fig:8B-Base_acc_all} further reports Pass@1 accuracy trajectories across ten mathematical reasoning benchmarks.
Across diverse datasets (e.g., AIME 2024/2025, OmniMath, and GPQA Diamond), TreeAdv-GRPO consistently converges faster and reaches a higher final accuracy than the GRPO baseline.

\subsection{Training Dynamics and Inference Latency Analysis}
\label{app:train-dyn-latency}

\begin{figure*}[!t]
    \centering
    \includegraphics[width=0.95\linewidth]{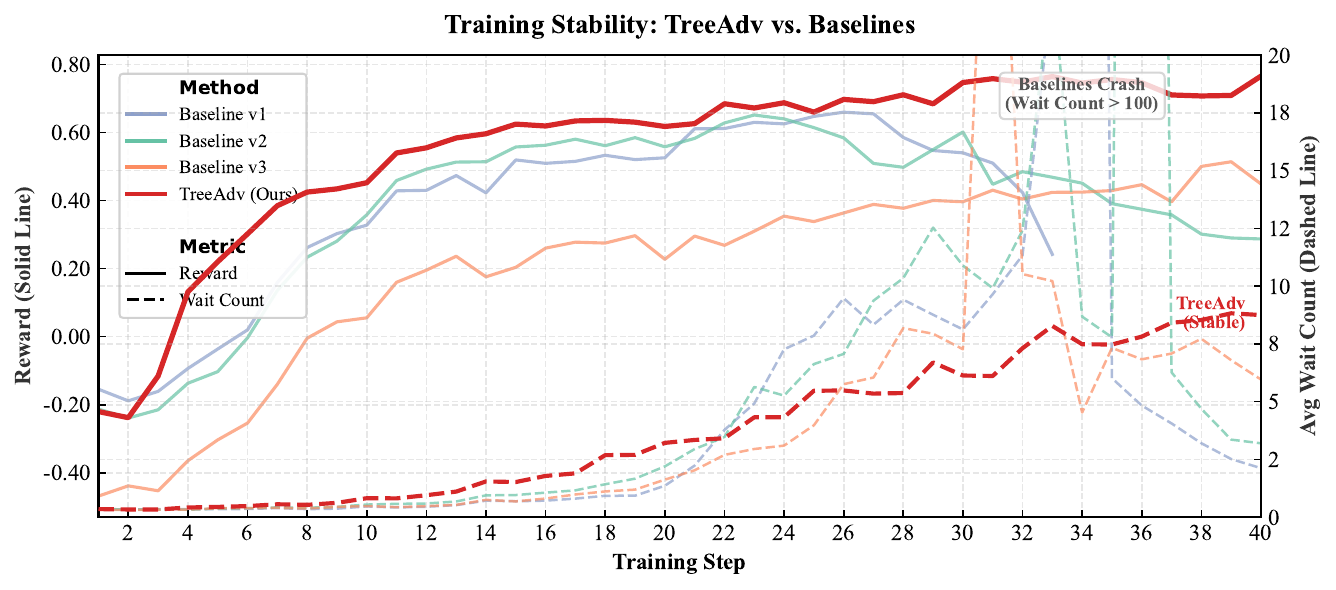}
    \caption{
    Comparison of training stability on Qwen3-8B-Instruct (Non-thinking mode). Solid lines represent the reward (left axis), while dashed lines denote the average wait count (right axis). All three baseline runs eventually suffer from a "wait count explosion" (exceeding 100, truncated here for visibility), leading to model collapse. In contrast, TreeAdv maintains robust stability with a consistently low wait count throughout the training process.
    }
    \label{fig:Dual_Axis_Reward_Wait}
\end{figure*}

In this section, we provide a detailed analysis of the training stability of \texttt{Qwen3-8B-Instruct} in the Non-thinking Mode. To ensure the reliability of our observations, we conducted three independent runs for the baseline GRPO. Figure~\ref{fig:Dual_Axis_Reward_Wait} illustrates the dual-axis comparison between these baseline runs and our proposed \textbf{TreeAdv-GRPO}, tracking both the training reward (solid lines, left axis) and the average wait count (dashed lines, right axis).

\paragraph{Consistent Collapse of Baseline GRPO.}
A critical and reproducible observation is the severe instability inherent in the baseline GRPO. As shown in Figure~\ref{fig:Dual_Axis_Reward_Wait}, all three baseline runs exhibit a catastrophic collapse in the later stages of training. We observe a strong negative correlation where the degradation in training reward coincides perfectly with a sudden explosion in inference latency.

Specifically, the average wait count for the baseline models spikes drastically, reaching values exceeding 100 (peaking at $\approx 137$ in one run). This significantly surpasses the visualization limit of the y-axis (truncated at 20 for clarity). This phenomenon suggests that without proper regularization, the baseline model falls into degenerate behaviors—generating excessively long, repetitive, or meaningless reasoning chains (``over-thinking'' without logic)—which consumes substantial computational resources while actively harming answer correctness.

\paragraph{Stability of TreeAdv-GRPO.}
In sharp contrast, \textbf{TreeAdv-GRPO} (red) demonstrates superior robustness under the exact same experimental settings. It effectively suppresses the tendency for latency explosion, maintaining the average wait count consistently below 10 throughout the entire training process. This indicates that our method successfully regularizes the model's reasoning path, preventing the emergence of inefficient long-context patterns and ensuring steady reward growth without the risk of mode collapse observed in the baselines.


\end{document}